# Edge Detection Based Shape Identification

Vivek Kumar#[1], Sumit Pandey#, Amrindra Pal#, Sandeep Sharma#
*# Department of Electronics and Communication Engineering*
*Dehradun Institute of Technology*
*Dehradun: 248009, India*
*[1]Student Author - er_vivek@outlook..com*

**Abstract: Image recognition is the need of the hour. In order to be able to recognize an image it is of immense importance that the image should be distinguishable from the background. In the present work, an approach is presented for automatic detection and recognition of regular two dimensional shapes in low noise environments. The work has large number of direct applications in the real world. The algorithm proposed is based on locating the edges and thus in turn calculating the area of the object helps in identification of a specified shape. The results were simulated using MATLAB tool are encouraging and validate the proposed algorithm.**

## I. INTRODUCTION

In recent years, edge detection technology has gradually been widely used in medicine, forestry, remote sensing, engineering, inspection of components, fault diagnosis and testing, and more. However, in practice, how to choose the right, a better image edge detection image processing and analysis is the focus of the study ([8]). Detection of regular geometric features like triangles, squares, rectangles and polygons in digital images is an important practice in image analysis and computer vision applications such as automatic inspection and assembly. Various methods for regular geometric shape detection have been researched till date [2]. The Hough Transform (HT) technique, one of the most popular one, has been used extensively to extract analytic features, such as straight lines, circles, and ellipse because of its robustness against noise, clutter, object defect, shape distortion etc. ([3], [4], [5], [6]).

On the flip side, algorithm like *Artificial Neural Network* is deployed in the recognition of various regular shapes. These algorithms can be used to perform nonlinear statistical modeling and provide a new alternative to logistic regression, the most commonly used method for developing predictive models for dichotomous outcomes in medicine. Neural networks offer a number of advantages, including requiring less formal statistical training, ability to implicitly detect complex nonlinear relationships between dependent and independent variables, ability to detect all possible interactions between predictor variables, and the availability of multiple training algorithms. However, due to the greater computational burden, proneness to over fitting, and the empirical nature of model development, these algorithms came to have some disadvantages. To figure out these loopholes and to overcome them a new algorithm is designed in the field of object recognition. *In this paper*, we focus on detecting certain two dimensional shapes with a newly approach based on the corner points identification and the area of the shape with less computing time and memory requirement than an artificial neural network does.

The paper has been sub-divided into seven sections. We premise our approach in the section II. Section III and IV caters light on the algorithm and the pseudo code regarding the approach used. Simulation results are provided in section V. Proceeding ahead, section VI derives the conclusion. The paper ends in the section VII giving the references of the work.

## II. PRINCIPLE OPERATION

The ultimate aim in a large number of image processing applications (like character recognition, medical image analysis, industrial automation, robotics, cartography, forensics, radar imaging, remote sensing etc. ) is to extract important *features* from image data, from which a description, interpretation, or understanding of the scene can be provided by the machine. For example, such as size and number of holes. More sophisticated vision systems are able to interpret the results of analyses and describe the various objects and their relationships in the scene. In this sense image analysis is quite different from the other image processing operations, such as restoration, enhancement, and coding where the output is another image. Image analysis basically involves the study of *feature extraction*, *segmentation*, and *classification* techniques ([10]) as depicted in Fig. 1.

It is a common problem in computer vision applications that regular or man-made objects should be recognized ([11], [12]) and so is the need for the recognition of irregular shapes. However, shape recognition of an irregular object can better be achieved if are able to recognize the most commonly seen regular shapes. Shape is commonly defined in terms of the set of contours that describe the boundary of an object. In contrast to gradient- and texture-based representations, shape is more descriptive at a larger scale, ideally capturing the object of interest as a whole. This has been recognized by the Gestalt school of perception, which has established the principle of holism in visual perception (Palmer, 1999; Ko-ka, 1935) ([9]).

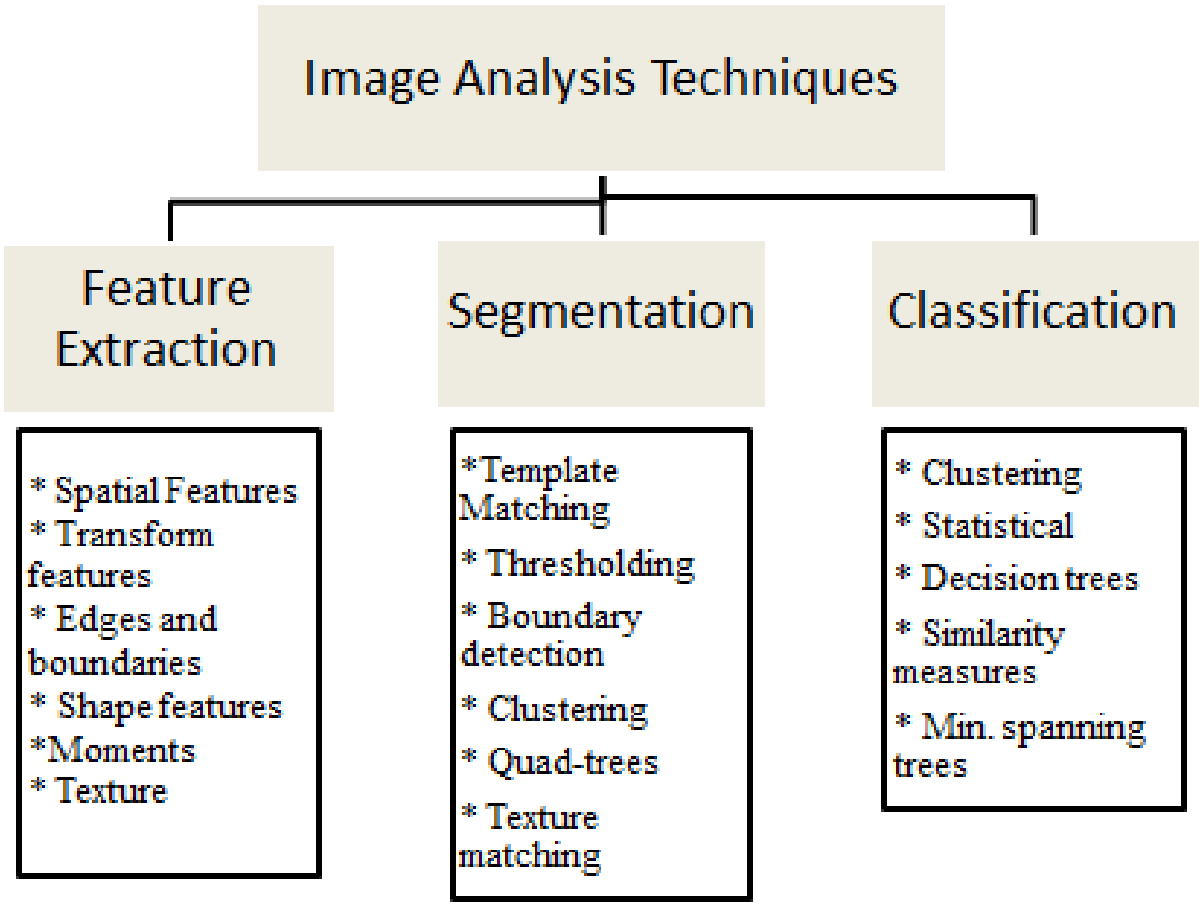

Fig. 1: Image Analysis Techniques

*Procedure:* When a RGB/Binary image is provided as an input to the MATLAB, few morphological operations are performed onto the image; next it is converted to grey scale or binary image. After the conversion the shape is chosen from the image to be detected and then its corners are marked by the four specified points (either the shape is a triangle, cone, cylinder etc. ). Moving forward, the distance is calculated among these given points and thence the area of the desired shape. Once, we have the distances and the area of the shape to be identified we can move to next step which classifies different conditions, on the basis of which shape can be distinguished among the given specified shapes:

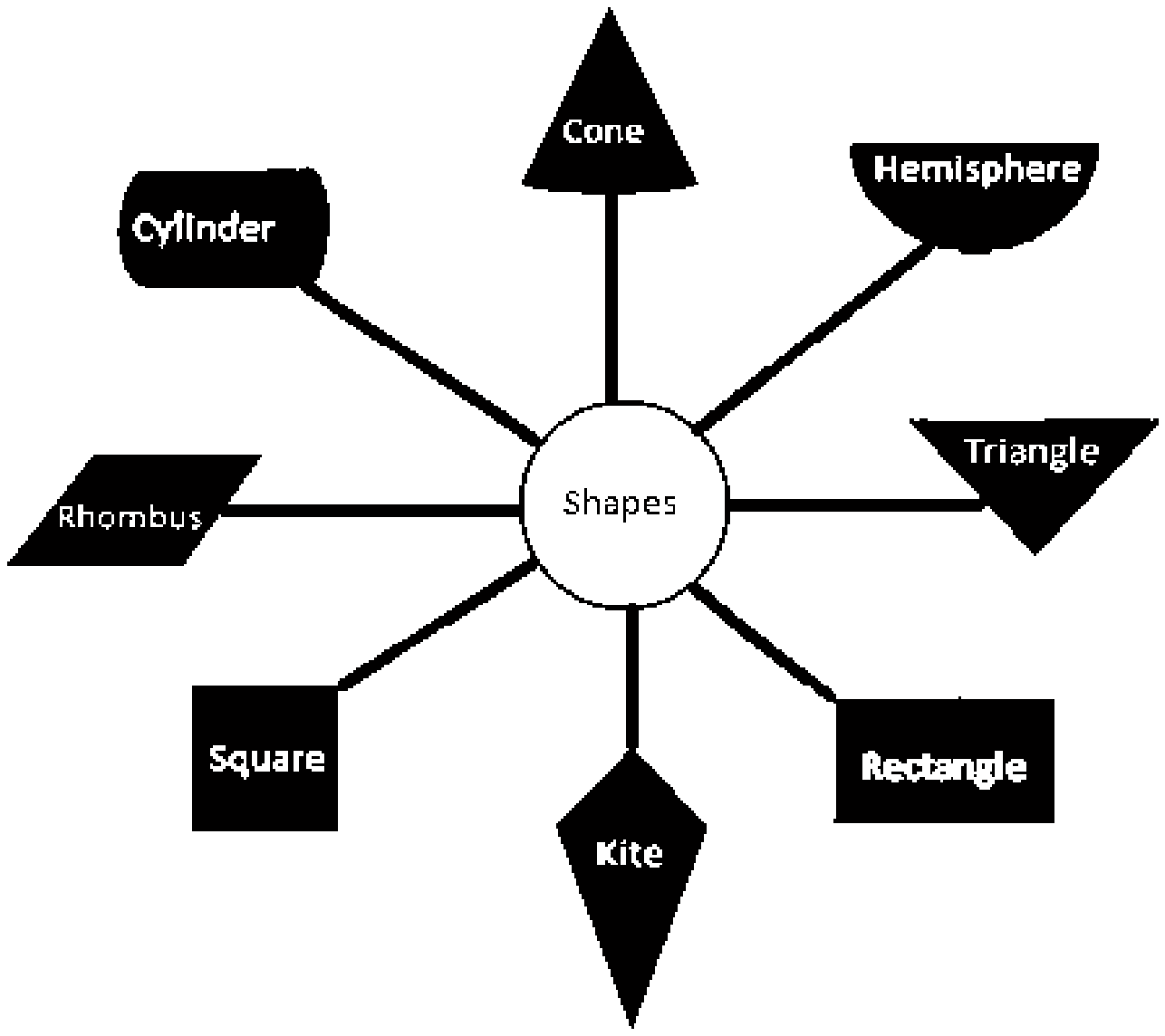

Fig. 2: Detectable 2-D Shapes

## III. ALGORITHM FOR CALCULATING DISTANCES AND AREA

i) *Calculation of Distances among four pre-defined points*:

Initially, four points have been marked on the corners of a shape to be detected by the algorithm itself. Now, the next step is to calculate the length of six distances among the four points so marked. These distances can be calculated by utilizing the formula specified in equation (1) in which ($x_1$, $y_1$) and ($x_2$, $y_2$) are the coordinates of any two points.

$$D = \sqrt{(x_2 - x_1)} + (y_2 - y_1) \qquad (1)$$

ii) *Calculating area by neighboring pixel in image:*

The area of a gray image can be obtained by counting on pixels in an image by summing the areas of each pixel in the image. The area of an individual pixel is determined by looking at its neighborhood. This technique is used for the surface area calculation of the image. In this area of computation process brighter pixels are counted.

a) Patterns with zero on pixel (area=0)
b) Patterns with one on pixel (area=1/4)
c) Patterns with two adjacent on pixels (area=1/2)
d) Patterns with two adjacent on pixels (area=3/4)
e) Patterns with three on pixels (area=7/8)
f) Patterns with all four on pixels (area=1)
This technique is simple and easy to obtain the area, and not affected by noise ([1]).

## IV. PSEUDO CODE REPRESENTATION FOR THE DETERMINATION OF THE IDENTITY OF A SHAPE

Each shape has a different *feature* and its own characteristics and unique-ness that separate it apart from another shape. Let $D_i$ ($i$ =1, 2, 3, 4, 5, 6) represents the six distances among the four points which are to be calculated. Furthermore, the pseudo codes for every shape are given below:

*a) Rectangle, Cylinder and Kite:*

```
for  i = 1 : 6
    if any of the two adjacent sides are equal to their
      opposite sides with a known permissible error
    then
        if  another two sides (i.e., diagonals) are equal
          with a known permissible error
        then
            if area = = 2πrh + 2πr^2 with known
              permissible error
                        disp (CYLINDER)
            else
                        disp (RECTANGLE)
            end
        else
                        disp (KITE)
        end
    end
end
```

*b) Square and Rhombus*

*for* i = 1 : 6
*if* any of the four sides are equal with a known permissible error
*then*
*if* another two sides (i.e., diagonals) are equal with a known permissible error
*then*
*disp (*SQUARE*)*
*else*
*disp (*RHOMBUS*)*
*end*
*end*
*end*

c) *Hemisphere*

Calculation of the radius by locating the center of the hemisphere which is the mid-point of the two same aligned horizontal/vertical points

*if* area =½ $\pi r^2$ with known permissible error
*disp(*HEMISPHERE*)*
*end*

*d) Triangle and Cone*

*if* sd < known permissible error && none other length = sd
(where sd = smallest distance among all distances)
*if* area == $\pi rl + \pi r^2$ with known permissible error
*disp (*CONE*)*
*else*
*disp (*TRIANGLE*)*
*end*
*end*

## V. SIMULATION RESULTS

We test our result on every shape to be detectable by proposed algorithm and measure the speed and accuracy of given approach so developed. Given below are the Original images and four point grey images. Four point grey images are broadened than original images just to bring the clarity for the *four red marked points* at the corners of each shape respectively.

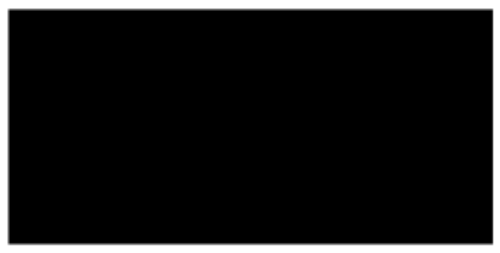

(a)

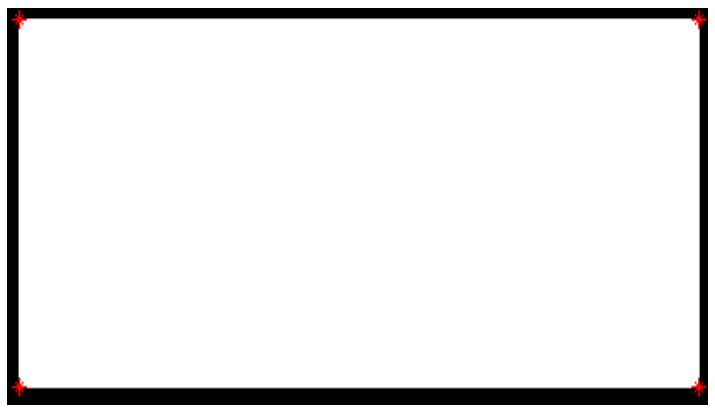

(b)

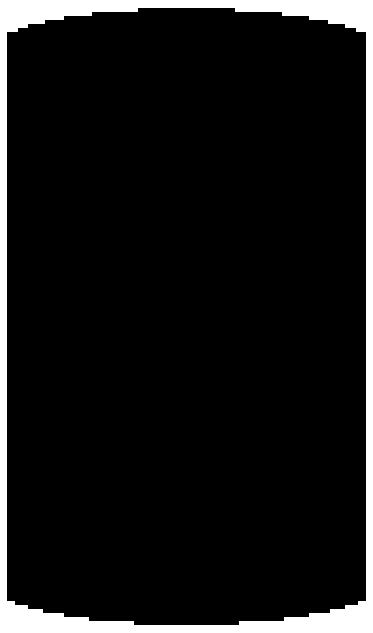

(c)

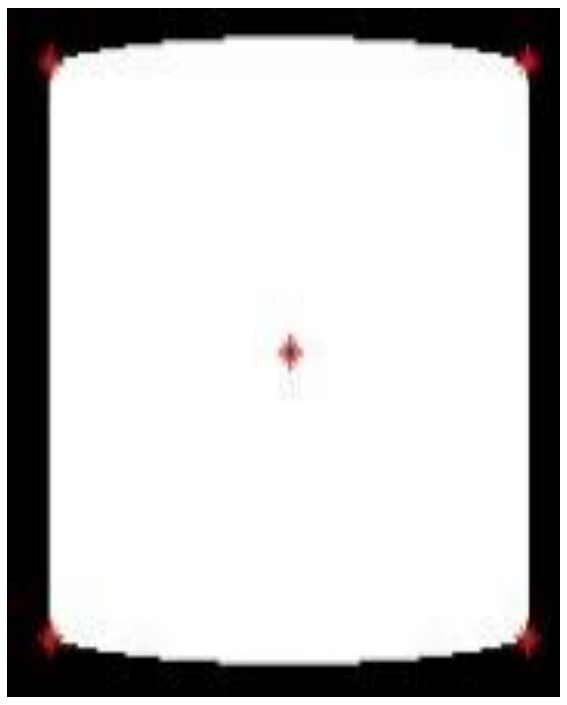

(d)

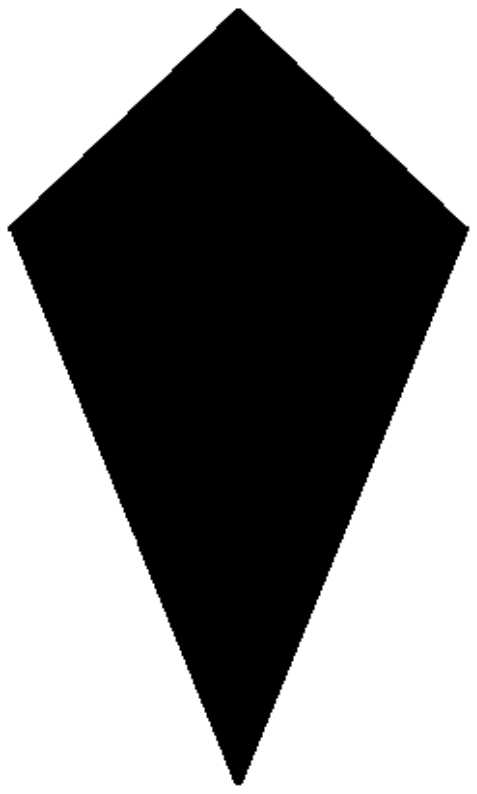

(e)

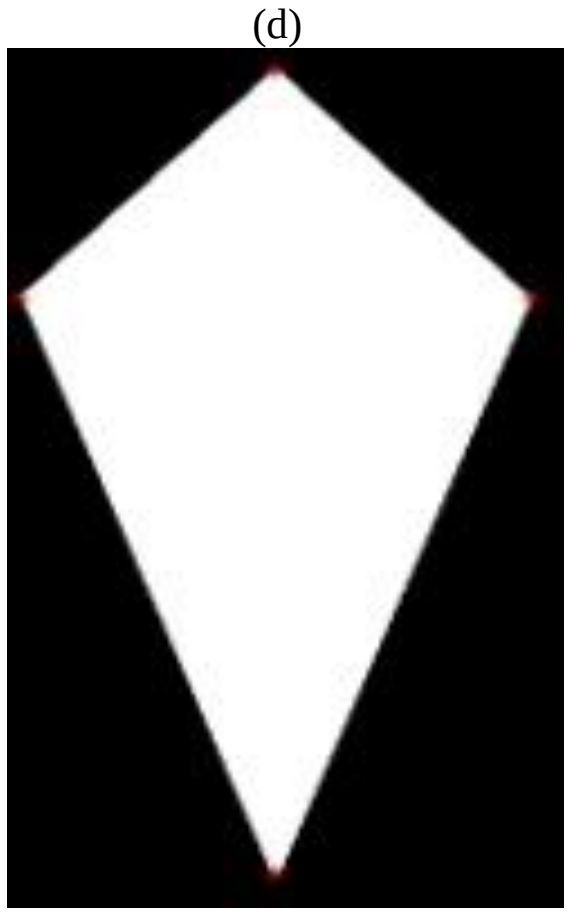

(f)

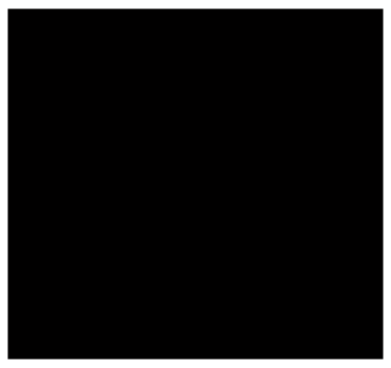

(g)

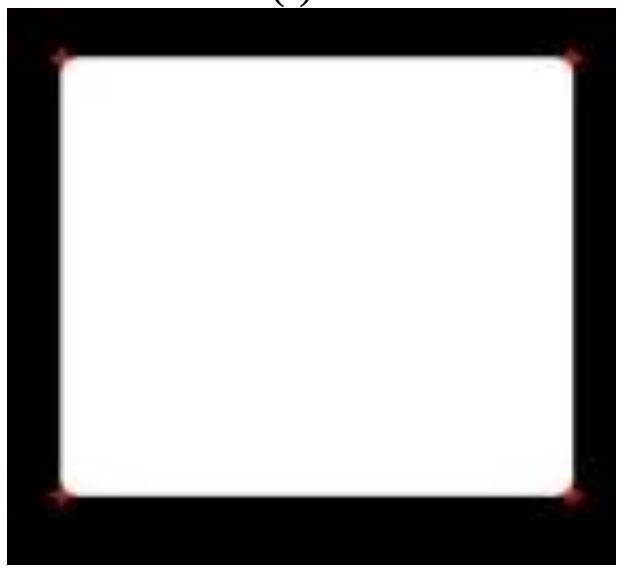

(h)

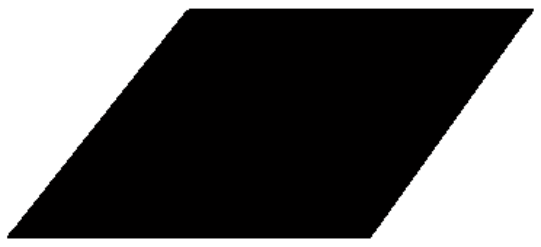

(i)

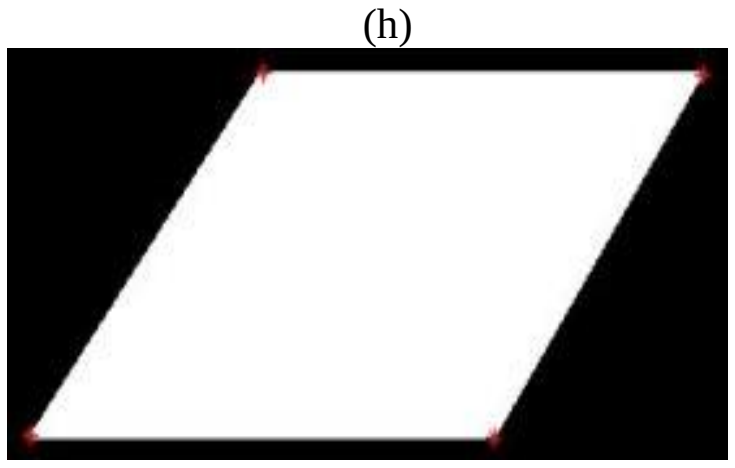

(j)

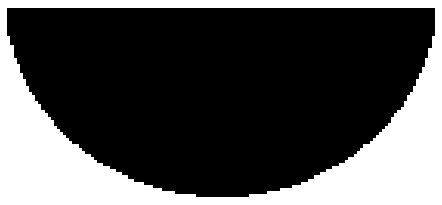

(k)

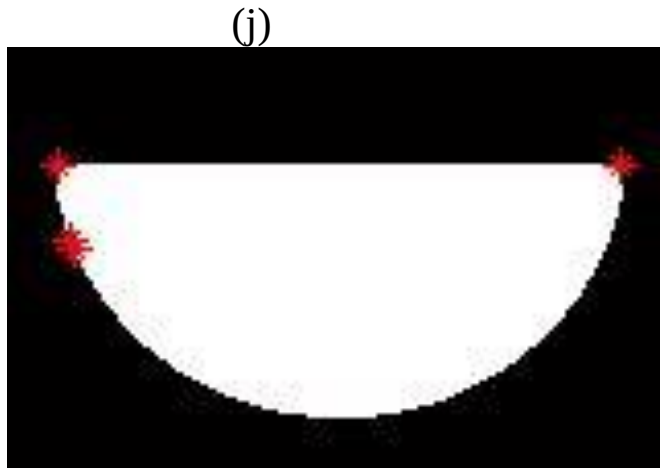

(l)

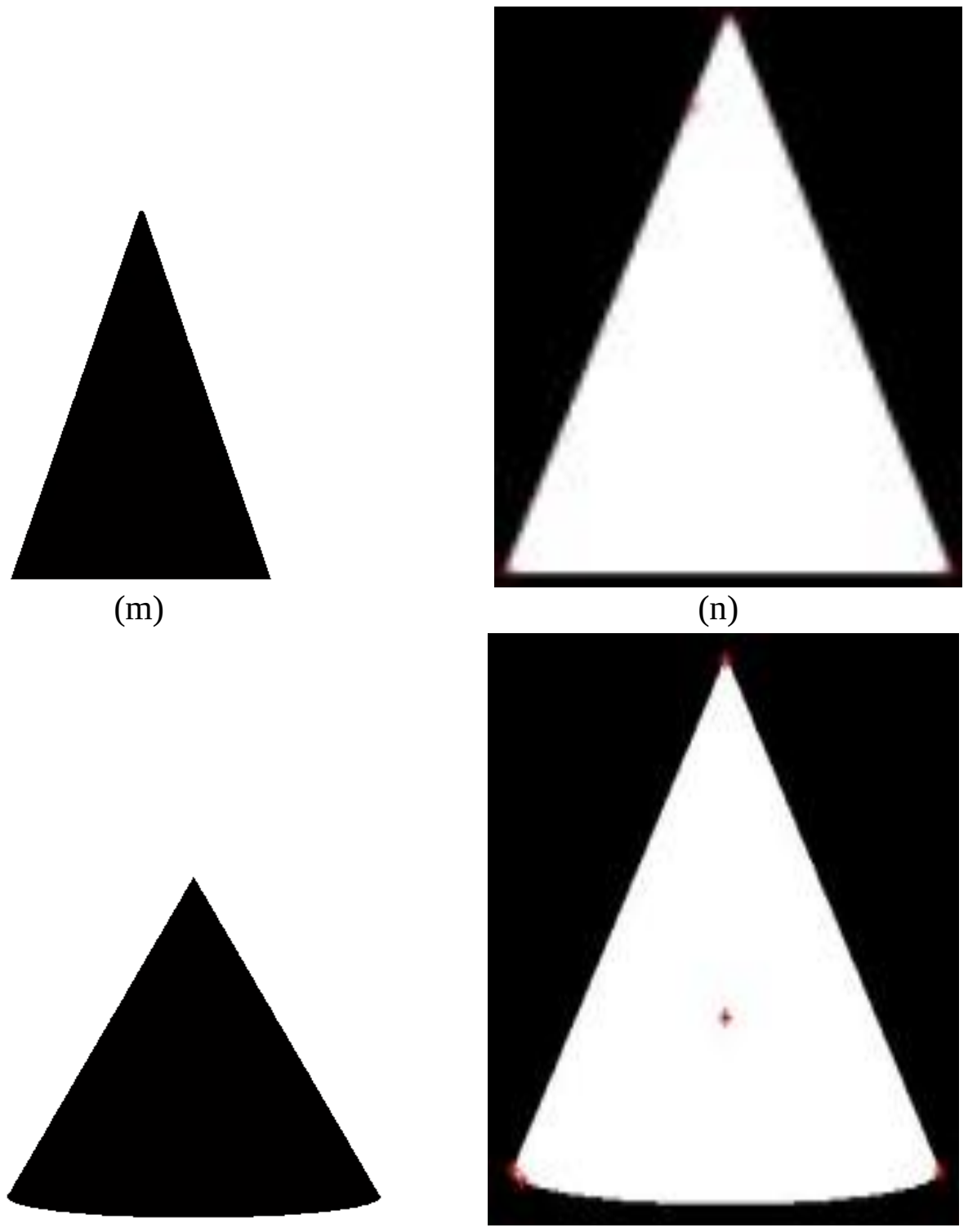

(m) (n)

(o) (p)

Fig. 1.2: (a) Rectangle* (b) Rectangle$^{\infty}$ (c) Cylinder* (d) Cylinder$^{\infty}$ (e) Kite* (f) Kite$^{\infty}$ (g) Square* (h) Square$^{\infty}$ (i) Rhombus* (j) Rhombus$^{\infty}$ (k) Hemisphere* (l) Hemisphere$^{\infty}$ (m) Triangle* (n) Triangle$^{\infty}$ (o) Cone* (p) Cone$^{\infty}$
where, * represents the original image, &
$^{\infty}$ represents the four points grey image.

The execution time corresponding to each shape is depicted in the following table:

| S. No. | Shapes | Execution Time (in sec) |
|---|---|---|
| 1. | Rectangle | 0.74084 |
| 2. | Cylinder | 0.85901 |
| 3. | Kite | 0.59109 |
| 4. | Square | 0.50946 |
| 5. | Rhombus | 0.47690 |
| 6. | Hemisphere | 0.45122 |
| 7. | Triangle | 0.47229 |
| 8. | Cone | 0.99982 |

## VI. CONCLUSION

The major advantage of this algorithm is that it can secernate various shapes and assures the correct result. This algorithm is time efficient as it requires just 0.6375 seconds on an average for executing the program and predicting the result. Not just favoring time efficiency, this algorithm is memory efficient too. Memory is used to store

(i) The coordinates of the pixel corresponding to each discovered corners.
(ii) The distance among the four points located on each corner respectively.
(iii) The area of the shape calculated by each neighboring brighter pixels.

Due to the above advantages, this algorithm is far most useful than a time taking and computational model like *Artificial Neural Network* to secernate various shapes. But the time complexity and memory requirement of our scheme have not been compared with other existing methods such as Hough Transform or GA based method ([7]). We are focusing on to comparing the proposed algorithm with those mentioned on the basis of parameters like memory usage, execution complexity, computing time and accuracy of system output in presence of noise ([2]).